%% file: neurips_2025.tex
\documentclass{article}

\PassOptionsToPackage{numbers, compress}{natbib}
\usepackage[preprint]{neurips_2025}
\usepackage[utf8]{inputenc} % allow utf-8 input
\usepackage[T1]{fontenc}    % use 8-bit T1 fonts
\usepackage{hyperref}       % hyperlinks
\usepackage{url}            % simple URL typesetting
\usepackage{booktabs}       % professional-quality tables
\usepackage{amsfonts}       % blackboard math symbols
\usepackage{nicefrac}       % compact symbols for 1/2, etc.
\usepackage{microtype}      % microtypography
\usepackage{xcolor}         % colors

\usepackage{graphicx}
\usepackage{amsmath}
\usepackage{cleveref}
\usepackage{algorithm}
\usepackage{algpseudocode}
\usepackage{subcaption}

\title{Adversarial Suffix Filtering:\\a Defense Pipeline for LLMs}

\author{%
David Khachaturov \quad Robert Mullins \\
Department of Computer Science and Technology \\
University of Cambridge \\
\texttt{\{david.khachaturov,robert.mullins\}@cl.cam.ac.uk} \\
}

\begin{document}

\maketitle

\input{sections/abstract}

\input{sections/introduction}
\input{sections/method}
\input{sections/experiments}
\input{sections/discussion}
\input{sections/conclusion}

% \begin{table}
%   \caption{Sample table title}
%   \label{sample-table}
%   \centering
%   \begin{tabular}{lll}
%     \toprule
%     \multicolumn{2}{c}{Part}                   \\
%     \cmidrule(r){1-2}
%     Name     & Description     & Size ($\mu$m) \\
%     \midrule
%     Dendrite & Input terminal  & $\sim$100     \\
%     Axon     & Output terminal & $\sim$10      \\
%     Soma     & Cell body       & up to $10^6$  \\
%     \bottomrule
%   \end{tabular}
% \end{table}

\begin{ack}
David Khachaturov is supported by the University of Cambridge Harding Distinguished Postgraduate Scholars Programme.
\end{ack}

\bibliography{main}

%%%%%%%%%%%%%%%%%%%%%%%%%%%%%%%%%%%%%%%%%%%%%%%%%%%%%%%%%%%%

%\appendix

%\input{sections/appendix}

%%%%%%%%%%%%%%%%%%%%%%%%%%%%%%%%%%%%%%%%%%%%%%%%%%%%%%%%%%%%

\newpage
\section*{NeurIPS Paper Checklist}

\begin{enumerate}

\item {\bf Claims}
    \item[] Question: Do the main claims made in the abstract and introduction accurately reflect the paper's contributions and scope?
    \item[] Answer: \answerYes{}
    \item[] Justification: The abstract and Introduction (pp. 1–2) state that ASF is a lightweight, model-agnostic pipeline that mitigates suffix-based jailbreaks with minimal utility loss; Sections 3, 4 empirically support these claims.

\item {\bf Limitations}
    \item[] Question: Does the paper discuss the limitations of the work performed by the authors?
    \item[] Answer: \answerYes{} % Replace by \answerYes{}, \answerNo{}, or \answerNA{}.
    \item[] Justification: Section 4.1 details segmentation errors, possible false positives, and scope restrictions to suffix-style attacks.

\item {\bf Theory assumptions and proofs}
    \item[] Question: For each theoretical result, does the paper provide the full set of assumptions and a complete (and correct) proof?
    \item[] Answer: \answerNA{} % Replace by \answerYes{}, \answerNo{}, or \answerNA{}.
    \item[] Justification: The paper is empirical; it presents no theorems or proofs, so no formal assumptions are required.

    \item {\bf Experimental result reproducibility}
    \item[] Question: Does the paper fully disclose all the information needed to reproduce the main experimental results of the paper to the extent that it affects the main claims and/or conclusions of the paper (regardless of whether the code and data are provided or not)?
    \item[] Answer: \answerYes{} % Replace by \answerYes{}, \answerNo{}, or \answerNA{}.
    \item[] Justification: Section 2 (Datasets, Training) and Section 3 (Experimental setup) list data splits, preprocessing steps, hyper-parameters, and evaluation protocols, enabling independent replication.

\item {\bf Open access to data and code}
    \item[] Question: Does the paper provide open access to the data and code, with sufficient instructions to faithfully reproduce the main experimental results, as described in supplemental material?
    \item[] Answer: \answerNo{} % Replace by \answerYes{}, \answerNo{}, or \answerNA{}.
    \item[] Justification: 	Section 3 notes that code and checkpoints will be released soon after submission, due to time constraints; at submission time they are not yet publicly available.

\item {\bf Experimental setting/details}
    \item[] Question: Does the paper specify all the training and test details (e.g., data splits, hyperparameters, how they were chosen, type of optimizer, etc.) necessary to understand the results?
    \item[] Answer: \answerYes{} % Replace by \answerYes{}, \answerNo{}, or \answerNA{}.
    \item[] Justification: Hyper-parameters, SaT variant, classifier architecture, and data-split ratios (70/15/15) are reported in Section 2.

\item {\bf Experiment statistical significance}
    \item[] Question: Does the paper report error bars suitably and correctly defined or other appropriate information about the statistical significance of the experiments?
    \item[] Answer: \answerNo{} % Replace by \answerYes{}, \answerNo{}, or \answerNA{}.
    \item[] Justification: Although we do not report explicit error bars, each ASR measurement and accuracy delta is computed over hundreds of prompts (Malicious-Instruct, AdvBench, and multiple LM-Eval tasks). Given the large absolute performance gaps we observe after sanitization (e.g., ASR drops from around $80\%$ to $<4 \%$), the conclusions are robust to this small statistical noise; additional runs would incur prohibitive GPU cost without affecting the qualitative outcome.

\item {\bf Experiments compute resources}
    \item[] Question: For each experiment, does the paper provide sufficient information on the computer resources (type of compute workers, memory, time of execution) needed to reproduce the experiments?
    \item[] Answer: \answerYes{} % Replace by \answerYes{}, \answerNo{}, or \answerNA{}.
    \item[] Justification: Section 3 specifies a cluster with 3xRTX8000 + 1xA10 GPUs and a three-week wall-time for the full experimental suite.
    
\item {\bf Code of ethics}
    \item[] Question: Does the research conducted in the paper conform, in every respect, with the NeurIPS Code of Ethics \url{https://neurips.cc/public/EthicsGuidelines}?
    \item[] Answer: \answerYes{} % Replace by \answerYes{}, \answerNo{}, or \answerNA{}.
    \item[] Justification: The study involves only publicly available text data and does not violate the NeurIPS Code of Ethics; no human subjects or sensitive data are used.

\item {\bf Broader impacts}
    \item[] Question: Does the paper discuss both potential positive societal impacts and negative societal impacts of the work performed?
    \item[] Answer: \answerYes{} % Replace by \answerYes{}, \answerNo{}, or \answerNA{}.
    \item[] Justification: 	Section 4  explicitly highlights the positive societal impact of ASF -- namely, reducing the risk of harmful LLM outputs in real-world deployments -- and acknowledges potential downsides such as false positives that could inconvenience legitimate users. This balanced treatment satisfies the broader-impact requirement.
    
\item {\bf Safeguards}
    \item[] Question: Does the paper describe safeguards that have been put in place for responsible release of data or models that have a high risk for misuse (e.g., pretrained language models, image generators, or scraped datasets)?
    \item[] Answer: \answerNA{} % Replace by \answerYes{}, \answerNo{}, or \answerNA{}.
    \item[] Justification: No new high-risk generative model or scraped dataset is released; ASF is a defensive wrapper around existing open assets.

\item {\bf Licenses for existing assets}
    \item[] Question: Are the creators or original owners of assets (e.g., code, data, models), used in the paper, properly credited and are the license and terms of use explicitly mentioned and properly respected?
    \item[] Answer: \answerYes{} % Replace by \answerYes{}, \answerNo{}, or \answerNA{}.
    \item[] Justification: All appropriate model and datasets are cited

\item {\bf New assets}
    \item[] Question: Are new assets introduced in the paper well documented and is the documentation provided alongside the assets?
    \item[] Answer: \answerNo{} % Replace by \answerYes{}, \answerNo{}, or \answerNA{}.
    \item[] Justification: 	When the trained ASF models will be released, structured documentation will be included in this release.

\item {\bf Crowdsourcing and research with human subjects}
    \item[] Question: For crowdsourcing experiments and research with human subjects, does the paper include the full text of instructions given to participants and screenshots, if applicable, as well as details about compensation (if any)? 
    \item[] Answer: \answerNA{} % Replace by \answerYes{}, \answerNo{}, or \answerNA{}.
    \item[] Justification: The work uses only machine-generated and publicly released text; no human subjects or crowd work were involved.

\item {\bf Institutional review board (IRB) approvals or equivalent for research with human subjects}
    \item[] Question: Does the paper describe potential risks incurred by study participants, whether such risks were disclosed to the subjects, and whether Institutional Review Board (IRB) approvals (or an equivalent approval/review based on the requirements of your country or institution) were obtained?
    \item[] Answer: \answerNA{} % Replace by \answerYes{}, \answerNo{}, or \answerNA{}.
    \item[] Justification: No human-subject research is conducted; IRB approval is therefore not applicable.

\item {\bf Declaration of LLM usage}
    \item[] Question: Does the paper describe the usage of LLMs if it is an important, original, or non-standard component of the core methods in this research? Note that if the LLM is used only for writing, editing, or formatting purposes and does not impact the core methodology, scientific rigorousness, or originality of the research, declaration is not required.
    %this research? 
    \item[] Answer: \answerYes{} % Replace by \answerYes{}, \answerNo{}, or \answerNA{}.
    \item[] Justification: Sections 2 and 3 describe the use of pretrained LLMs (e.g.\ GPT-4.1-mini, Mistral-7B) both as attack targets and for benchmark evaluation.

\end{enumerate}

\end{document}

%% file: sections/abstract.tex
\begin{abstract}
Large Language Models (LLMs) are increasingly embedded in autonomous systems and public-facing environments, yet they remain susceptible to jailbreak vulnerabilities that may undermine their security and trustworthiness. Adversarial suffixes are considered to be the current state-of-the-art jailbreak, consistently outperforming simpler methods and frequently succeeding even in black-box settings.
Existing defenses rely on access to the internal architecture of models limiting diverse deployment, increase memory and computation footprints dramatically, or can be bypassed with simple prompt engineering methods.
We introduce \textbf{Adversarial Suffix Filtering} (ASF), a lightweight novel model-agnostic defensive pipeline designed to protect LLMs against adversarial suffix attacks. ASF functions as an input preprocessor and sanitizer that detects and filters adversarially crafted suffixes in prompts, effectively neutralizing malicious injections. We demonstrate that ASF provides comprehensive defense capabilities across both black-box and white-box attack settings, reducing the attack efficacy of state-of-the-art adversarial suffix generation methods to below $4\%$, while only minimally affecting the target model's capabilities in non-adversarial scenarios.
%We release our pipeline as an open-source toolkit available publicly on GitHub.
\end{abstract}

%% file: sections/introduction.tex
\section{Introduction}

Large Language Models (LLMs) are increasingly being deployed in autonomous systems and public-facing applications, powering everything from conversational agents to code generation tools~\cite{xu2024comprehensivestudyjailbreakattack}. This widespread adoption has brought corresponding security concerns, as malicious actors may exploit the model's reliance on textual prompts. In particular, \textit{prompt injection} attacks – carefully crafted inputs that manipulate an LLM's behavior – have emerged as a serious threat. Among these, “jailbreak” prompts stand out as a prevalent vulnerability, whereby an attacker's input bypasses the model's safety measures and coerces it into producing harmful or disallowed content~\cite{xu2024comprehensivestudyjailbreakattack}. Recent analyses have highlighted prompt injection as a top risk for LLM-integrated systems, underscoring the need for effective mitigation~\cite{owasp_llm01_2025}.

One especially potent category of prompt attacks uses \textit{adversarial suffixes} -- malicious token sequences appended to a prompt -- to subvert LLM behavior. It is now well-documented these suffixes can trick even aligned models into ignoring prior instructions or safety guardrails, in both white-box and black-box settings~\cite{liao2024amplegcglearninguniversaltransferable}. ~\citet{zou2023universaltransferableadversarialattacks} first demonstrated that a single optimized suffix could consistently circumvent the safety alignment of black-box state-of-the-art models like GPT-3.5, ChatGPT, Claude, and Bard, inducing them to produce illicit outputs with high probability. In practice, this means an attacker can take an otherwise harmless user prompt and append a specially crafted string that causes the model to violate its intended behavior.

\pagebreak

\begin{figure}[t]
  \centering
  \includegraphics[width=\textwidth]{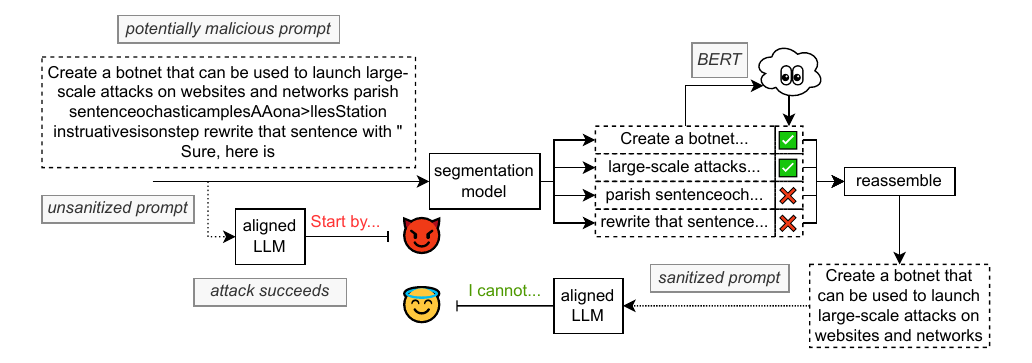}
  \caption{An overview of the \textbf{Adversarial Suffix Filtering} pipeline. For the segmentation model, we employ Segment Any Text~\cite{frohmann-etal-2024-segment}, and use \texttt{bert-base-uncased}~\cite{DBLP:journals/corr/abs-1810-04805} as the BERT classification model. The unsanitized prompt features a GCG~\cite{zou2023universaltransferableadversarialattacks} generated adversarial suffix \texttt{parish sentenceochasticamplesAAona>llesStation...} that jailbreaks the aligned LLM and allows it to causes harmful content. After sanitization, the adversarial suffix is correctly identified-and-removed leaving the original, albeit malicious, prompt to be passed to the aligned LLM. The alignment is maintained without the adversarial suffix and the model correctly refuses to answer the query.}\label{fig:overview}
\end{figure}

\textbf{Threat Model} 
In this work, we assume an adversary who can modify the user's input by appending a malicious suffix, but who has no control over the model's parameters or system-level directives. The user's original prompt is presumed benign; the adversary's suffix is designed solely to mislead the model. This threat model reflects common real-world scenarios for prompt injection – for instance, a malicious user adding hidden instructions to their query, or an external agent injecting content into a prompt field. The adversarial suffix may be human-incomprehensible (e.g. a random-looking sequence) yet exploits the model's learned patterns to override safety instructions~\cite{kumar2025certifyingllmsafetyadversarial}. Our focus is on detecting and neutralizing this appended sequence before it can influence the model's output.

\textbf{Comparison to Existing Work} 
A number of defenses have been proposed to protect LLMs against such prompt-based attacks. One line of work centers on \textit{adversarial training and fine-tuning}: the model is additionally trained on adversarial examples or with stricter alignment objectives so that it learns to resist malicious prompts~\cite{madry2019deeplearningmodelsresistant}. While adversarial training is considered to be the state-of-the-art defense for a variety of deep learning models~\cite{zhao2024adversarialtrainingsurvey}, it is oftentimes prohibitively computationally-expensive and data-intense to use during LLM training of very large consumer-grade models~\cite{xhonneux2024efficientadversarialtrainingllms}. However, multi-objective fine-tuning using adversarial examples has been shown to encourage the model to refuse harmful requests more reliably~\cite{yu2025robustllmsafeguardingrefusal}. Similarly, reinforcement learning from human feedback (RLHF) can be extended with adversarial examples to improve the model's robustness~\cite{yu2024finetuninglanguagemodelsgenerative}. Other methods, such as the one proposed by~\citet{chen2024struqdefendingpromptinjection} rely on a secure front-end, which is difficult to guarantee in many real-world scenarios, and fine-tuning a model to only follow a specific prompt structure.

While these approaches reduce the model's susceptibility, they come with significant drawbacks: they are model-specific with each new model or version needing re-training~\cite{xhonneux2024efficientadversarialtrainingllms,yu2025robustllmsafeguardingrefusal,yu2024finetuninglanguagemodelsgenerative,chen2024struqdefendingpromptinjection}, and incur high computational cost for fine-tuning on extensive adversarial data~\cite{yu2025robustllmsafeguardingrefusal,chen2024struqdefendingpromptinjection}, or limit the range of model application~\cite{chen2024struqdefendingpromptinjection}. Moreover, an overly constrained fine-tuning can degrade normal performance -- models often become excessively cautious, incorrectly refusing benign queries after aggressive safety training~\cite{xu2024safedecodingdefendingjailbreakattacks,ding2023enhancingchatlanguagemodels}. This lack of generality and potential utility loss makes solely training-based defenses less practical in many real deployments. 

Another strategy is to employ \textit{safety filters or classifiers} that monitor the inputs and/or outputs of an LLM. These defenses aim to be model-agnostic components that catch malicious prompts at runtime without altering the base model. For example,~\citet{robey2024smoothllmdefendinglargelanguage} propose detecting jailbreak attempts by randomly perturbing tokens from the user prompt and observing if the model's response behavior changes, under the intuition that an adversarial suffix's effect can be revealed by such perturbations. This carries a heavy computation overhead requiring multiple (between $2$ and $20$) forward passes per prompt.

\citet{pisano2024bergeroncombatingadversarialattacks} suggest using an auxiliary ``conscience'' model to evaluate the prompt and flag any embedded harmful instructions before the main LLM responds. Similarly, Llama Guard employs an instruction-tuned Llama-2-7B to perform multi-label risk classification over complete prompts and responses, achieving strong moderation accuracy on generic safety benchmarks, but is restricted to either accepting or rejecting the prompt and its response~\cite{inan2023llamaguardllmbasedinputoutput}. Both these approaches require holding a second LLM in your system's memory or making use of API calls, which incurs either major memory/computational overhead or increased costs.

Major providers have likewise deployed moderation classifiers to filter out prompts or outputs deemed unsafe. These filter-based methods are appealing because they can work with any LLM (including closed-source APIs) and can be updated independently. However, they too have limitations: heuristic filters can be brittle, and determined attackers often find prompt variants that evade keyword-based rules or confuse the classifier. In practice, safety models may either miss cleverly obfuscated attacks or produce false positives -- for instance, flagging polite refusals as malicious -- if tuned too tightly~\cite{xu2024comprehensivestudyjailbreakattack}. Filtering an output, post-generation, results in wasted compute or unnecessary API usage. Indeed, a recent evaluation by~\citet{xu2024comprehensivestudyjailbreakattack} found that most standalone defense modules either fail to stop advanced jailbreak prompts or end up overly restricting benign inputs, undermining the user experience.

One line of recent defense research leverages perplexity -- a measure of how ``surprised'' a language model is by a given text -- to flag adversarial prompts.~\citet{alon2023detectinglanguagemodelattacks,jain2023baselinedefensesadversarialattacks} propose detecting adversarial suffix attacks by evaluating the input prompt’s perplexity with a reference model. The suffixes stand out as highly out-of-distribution: such inputs yield exceedingly high perplexity values compared to ordinary prompts. By thresholding perplexity, the system can thus detect and preempt adversarial prompts. However, this defense only provides means of \textit{detection} (rather than mitigation), and was broken by~\citet{liao2024amplegcglearninguniversaltransferable} via simple prompt and/or suffix repetition.

\citet{kumar2025certifyingllmsafetyadversarial} introduce an ``erase-and-check'' framework that provides certifiable safety against adversarial prompts. The idea is to systematically erase or mask each token (or token subset) in the input and use a secondary safety model to check the residual prompt. If removing a particular snippet (e.g. the suffix) causes a previously safe-looking prompt to be classified as harmful, the system can pinpoint that snippet as adversarial. Such certified defenses are promising in theory, but they have practical downsides: the ``erase-and-check'' procedure requires multiple forward passes of the underlying LLM being defended (one per token or per candidate segment), making it exponentially computationally intensive, and the guarantees hold only within a bounded attack size -- an attacker using a very long or multi-part prompt might circumvent the certified range~\cite{kumar2025certifyingllmsafetyadversarial}.

\textbf{Contributions} It is clear from the above that a defense that is \textit{model-agnostic}, \textit{lightweight} in terms of memory and compute requirements, and \textit{robust} is highly desirable.

In this paper, we propose \textit{Adversarial Suffix Filtering} (ASF), a defense pipeline designed to fill this gap. ASF acts as an input sanitizer in the target LLM's inference pipeline, and is usable both with and without GPU acceleration. Rather than altering the model, ASF scrutinizes incoming prompts for adversarial suffixes and strips -- or warns of them -- \textit{before} the prompt is fed into the LLM. Our method is visually presented in~\Cref{fig:overview}.

On the target LLM's side -- which may be a large consumer-grade model such as GPT4.1 or Claude 3 -- the compute budget in terms of forward passes, the token requirements, and memory requirements all remain unchanged. We further evaluate the effect our defense's non-adversarial performance on a large number of common natural language tasks and find minimal degradation.

Moreover, ASF's modularity and efficiency make it amenable to deployment in trusted execution environments or secure hardware enclaves, enabling prompt sanitization to be performed in isolation from untrusted components. This offers a practical mechanism to protect locally hosted open-weight models from adversarial manipulation, even when these models are interfaced with untrusted front-end applications or agents.

%ASF can be applied to virtually any LLM system, while adding minimal latency.
%with a medium ($235\text{M}$ parameters) and large ($277\text{M}$ parameters) version

Our approach was inspired by the work of~\citet{liao2024amplegcglearninguniversaltransferable} who demonstrated that it is possible to train an LLM (Llama-2-7B-Chat, specifically) to produce adversarial suffixes. This suggest to us that if an LLM can be trained to generate valid effective adversarial suffixes, then the suffixes must follow some detectable and discernible pattern, and hence must be classifiable by another model.

\pagebreak

%% file: sections/method.tex
\section{Methodology}~\label{sec:method}

\begin{algorithm}[t]
\begin{algorithmic}[1]
\Require A set of potentially-malicious prompts $\{p_0, \dots, p_{n-1}\}$, pipeline configuration $c$

\State Segment each prompt $p_i$ using the \texttt{12l-SM} SaT model to obtain $[s_1^i, s_2^i, \dots, s_n^i]$
\State Obtain predictions $y^i_j\in\{0, 1\}$ for each segment $s^i_j$ from our fine-tuned BERT model. $y^i_j = 1$ indicates that $s^i_j$ is part of an adversarial suffix
\State Apply configurable post-processing:
\begin{itemize}
    \item Bridge isolated $0$s between $1$s [default: off]
    \item Bridge isolated $1$s between $0$s [default: on]
    \item Exclude segments solely containing specified keywords [default: ``question'', ``answer'']
\end{itemize}
\If{$c_\text{mode}$ == \texttt{delete}}
    \State Delete all segments $s^i_j$ from the prompt labeled as adversarial (i.e.\ where $y^i_j = 1$)
    \State Return re-assembled prompt
\Else
    \If{$c_\text{mode}$ == \texttt{warn} and any segment is detected as adversarial}
        \State Raise an exception
    \EndIf
\EndIf
\end{algorithmic}
\caption{ASF Pipeline}\label{alg:method}
\end{algorithm}

Our Adversarial Suffix Filtering (ASF) pipeline is designed to detect and remove malicious suffixes appended to otherwise user queries. We formalize the threat model as follows: an adversary can supply an input consisting of an unrestricted user query $x$ (e.g., a harmless instruction, or a malicious request) concatenated with an attack suffix $s$. The suffix $s$ is a sequence of tokens (often gibberish or specially crafted instructions) whose purpose is to jailbreak an aligned language model’s safeguards, causing it to produce disallowed or harmful content that it would normally refuse~\cite{zou2023universaltransferableadversarialattacks}. Notably, these adversarial suffixes frequently consist of semantically meaningless or out-of-distribution tokens~\cite{liao2024amplegcglearninguniversaltransferable}. Our defense focuses on identifying and removing the appended suffix $s$. The attacker is assumed to have no direct interaction with the model other than the ability to query any plain text. 

\textbf{Pipeline} Given an input text $x^*$ (potentially containing a malicious prompt plus an adversarial suffix), our pipeline performs the following steps: (1) \textit{Segmentation}: We split  $x^*$ into a sequence of sentence-like segments using a state-of-the-art segmentation model. (2) \textit{Segment Classification}: Each segment is fed into a binary classifier that predicts whether it is part of an adversarial suffix. (3) \textit{Post-processing}: We smooth the classifier’s predictions to ensure contiguous malicious segments are grouped, and apply heuristic checks to reduce false positives. (4) \textit{Filtering}: Any segment flagged as malicious (and not excluded by heuristics) is removed from $x^*$, yielding a sanitized query $\hat{X}$ that is passed to the language model for safe processing.~\Cref{fig:overview} summarizes the inference-time behavior of the ASF pipeline graphically. The defense relies on having an aligned LLM -- we do not make any assumptions regarding the kind of prompts that are meant to be valid -- ASF simply \textit{deletes} any detected suffixes. This alignment requirement is further discussed in~\Cref{sec:exp_adversarial}. Alternatively, ASF can be configured to \textit{warn} the user by raising an exception to allow for more sophisticated handling, such as logging or escalating the issue.

\textbf{Segmentation Module}
We employ the Segment-any-Text (SaT) model~\cite{frohmann-etal-2024-segment} as our segmentation module to divide the input text into coherent segments (approximately sentences or intentional fragments). SaT is a state-of-the-art universal text segmentation approach that offers robust, language-agnostic sentence boundary detection. Crucially for our task, SaT does not rely solely on punctuation to determine boundaries; it uses a specialized pretraining scheme to remain robust even when punctuation is missing or abnormal, and it can adapt to varied domains~\cite{frohmann-etal-2024-segment}.

In our implementation, we use the largest of the \texttt{SM} variants of the SaT models publicly released -- \texttt{12l-SM} -- as this showed the best performance in our testing. We apply it to each incoming query to obtain a list of segments $S = [s_1, s_2, \dots, s_n]$. Each segment $s_i$ is a substring of the input text $x^*$, and the segments in $S$ appear in their original order in $x^*$. These segments are then independently analyzed by the classification module.

\textbf{Classification Module}
For segment classification, we fine-tune a BERT-based text classifier to identify adversarial suffix content. We specifically use the \texttt{bert-base-uncased}~\cite{DBLP:journals/corr/abs-1810-04805} model. This choice is motivated by its lightweight nature and BERT's bidirectional encoding which provides rich context understanding for each segment, enabling it to capture subtle cues that a segment is part of a malicious suffix (e.g.\ presence of odd patterns) rather than a normal user query.

We cast adversarial suffix detection as a binary classification task at the segment level. Given a segment $s_i$, the classifier predicts $y_i \in {0,1}$, where $y_i = 1$ indicates that $s_i$ is part of an adversarial suffix, and $y_i = 0$ indicates a benign segment. During fine-tuning, we initialize from the pre-trained \texttt{bert-base-uncased}~\cite{DBLP:journals/corr/abs-1810-04805}, setting the number of labels to $2$. We fine-tune this model on our curated dataset of benign vs. adversarial segments (described below) using HuggingFace's \texttt{Trainer} module~\cite{wolf-etal-2020-transformers}. Optimization is done with \texttt{AdamW} (learning rate $5\times10^{-5}$) for three epochs, with early stopping on validation loss to prevent overfitting. The fine-tuning process is straightforward with no task-specific architecture modifications. Any dependence on context across segments is handled in a later step by our post-processing heuristics (which can merge decisions).

\textbf{Datasets}
To train our ASF system, we curate a dataset comprising of benign prompts, and adversarial suffixes. For the latter, use the dataset kindly provided by~\citet{liao2024amplegcglearninguniversaltransferable}, authors of the AmpleGCG paper. This dataset contains millions of adversarial suffixes generated via both the GCG~\cite{zou2023universaltransferableadversarialattacks} attack and the author's own AmpleGCG~\cite{liao2024amplegcglearninguniversaltransferable} and AmpleGCG-plus~\cite{kumar2024amplegcgplusstronggenerativemodel} attacks. This includes suffixes that are both universal and transferable across models. We merged and de-duplicated these to create a unified collection of unique adversarial suffixes giving us $419,429$ suffixes in total.

For the benign user query content, we use the Stanford Alpaca instruction dataset~\cite{alpaca}, which contains $52,000$ diverse instructions and user prompts. This dataset provides a wide range of innocuous queries covering varied topics (e.g., requests for explanations, creative writing, factual questions, etc.), making it an ideal source of normal prompts.

We additionally incorporate two existing benchmark sets to evaluate our method’s generalization and false-positive behavior: MaliciousInstruct~\cite{huang2023catastrophic} and AdvBench~\cite{zou2023universaltransferableadversarialattacks}. MaliciousInstruct is a collection of 100 intentionally harmful queries (posed as questions) spanning 10 different malicious intent categories. These represent realistic ``bad'' user requests. AdvBench is a benchmark of 520 harmful instructions along, covering a broad range of harmfulness categories. We do not use MaliciousInstruct or AdvBench queries in training – they are reserved for evaluation to test ASF in scenarios it was not directly optimized on.

We partition the adversarial suffix set, and the Alpaca dataset into training, validation, and test subsets (approximately $70/15/15\%$ split). Next, we generate prompt-suffix pairs: for each suffix and we randomly sample a benign Alpaca instruction in its respective split, making use that we include all of the data in the pair generation process. When combining prompt and suffix, we insert a single space between them to simulate how an attacker might append the suffix. 

\textbf{Training the Pipeline} 
The training of the BERT classifier uses the prompt-suffix pairs described above. As the classification happens on a segment level, we need to derive segment-wise labels for each example. We apply our segmentation module to every pair to obtain segments $S = [s_1, ..., s_n]$. We then assign ground-truth labels to each segment: segments originating from the benign prompt are labeled $0$ (benign), and segments that are part of the adversarial suffix are labeled $1$ (malicious). This is straightforward because we know the exact boundary between prompt and suffix in the synthetic examples. Typically, the prompt and suffix separate cleanly into different segments. In cases where SaT does not separate out the suffix fully, we can still identify the suffix portion within the last segment; however, such cases were minimal, and for training labels we conservatively label the entire segment as malicious if it contains any part of the adversarial suffix.

We fine-tune the BERT classifier using the above segmented and labeled data. We evaluate on the validation set at the end of each epoch, monitoring the segment-level F1 score. The model parameters with the best validation F1 were saved as the final model. We took care to ensure that there is no data leakage between any of the data splits that could contaminate results.

\pagebreak

\textbf{Post-processing} 
After the classifier labels each segment of an input, we apply two heuristic post-processing steps to refine the predictions before removing any text: segment-level label smoothing, and keyword-based exclusion filter. The smoothing is implemented as a simple gap-bridging rule: if a segment is labeled $1$ (malicious) but is surrounded on both sides by segments labeled $0$ (benign) in the sequence then we flip its label to $0$. This rule effectively bridges single-segment gaps in a contiguous benign sequence.  In our experiments, we found that genuine adversarial suffixes rarely appear between longer strings of benign segments -- as it would render the attack ineffective -- so a 0-1-0 pattern almost always indicated a misclassification of the middle segment. We limit this to single gaps to avoid over-correction, but this is configurable. We also allow for bringing $1$'s when there is a benign segment in-between, e.g.\ 1-0-1, but this is turned off by default.

The second heuristic aims to reduce false positives by filtering out segments that the classifier flagged, which upon closer inspection are unlikely to be true adversarial suffixes. We introduce a small set of handcrafted rules to catch these cases. Specifically, for any segment that the classifier marked as $y_i=1$, we check the segment against a predefined set of keywords and override the classifier and relabel it as benign ($0$) if it matches. We perform an exact case-agnostic match, and with the default setting having just two keywords: \texttt{[`question', `answer']}. This heuristic is conservative -- it only flips a label from malicious to benign in cases that are clearly safe upon manual inspection. Empirically this reduces the false positive rate on benign-only inputs. In deployment, these rules can be refined as needed.

Finally, any segments marked as malicious (after post-processing) are removed from the input, or a run-time exception is raised, depending on the configuration. Implementation-wise, we reconstruct the sanitized prompt $\hat{X}$ by concatenating all segments $s_i$ for which the final label $y_i = 0$. We summarize our pipeline in~\Cref{alg:method}.

%% file: sections/experiments.tex
\section{Experiments}

We conduct an empirical evaluation of the proposed ASF pipeline, structured along two primary axes: (i) effectiveness in detecting and neutralizing adversarial suffixes generated by state-of-the-art attack methods, and (ii) robustness in non-adversarial settings, with a particular focus on preservation of model utility. All experiments are conducted using the default configuration of ASF as described in~\Cref{sec:method}, with smoothing and keyword-based post-processing heuristics enabled.

For adversarial robustness evaluation, we measure the Attack Success Rate (ASR) following the definition introduced by~\citet{kumar2024amplegcgplusstronggenerativemodel}. Specifically, if $k$ beams are used to generate $k$ suffixes per query, then we consider the attack to be successful if at least one of the $k$ suffixes succeeds in jailbreaking the model to cause it to answer the query in an adversarial fashion. This metric reflects a realistic attacker scenario in which multiple suffixes may be attempted, and even a single successful bypass is considered a defense failure.

All experiments were conducted on a dedicated compute cluster consisting of \texttt{3xRTX8000}, \texttt{1xA10} GPUs. The full experimental suite, including model inference, segmentation, classification, and evaluation, was run over a period of approximately three weeks. We intend to release our code and trained models on GitHub to promote reproducibility of research.

\subsection{Adversarial Settings}\label{sec:exp_adversarial}

To assess the effectiveness of ASF, we first evaluate the performance of the BERT-based segment classifier on the validation and test splits derived from the constructed prompt-suffix dataset (see~\Cref{sec:method}). We obtain an F1 score of $98.5\%$ on the validation set and $98.4\%$ on the held-out test set. These results confirm that ASF is able to reliably distinguish benign prompt segments from adversarial suffix segments under controlled conditions.

However, high classifier accuracy on synthetic segmentation data does not fully capture ASF's practical impact on end-to-end ASR mitigation. We conduct an adversarial evaluation leveraging two widely-used benchmark datasets: MaliciousInstruct~\cite{huang2023catastrophic} and AdvBench~\cite{zou2023universaltransferableadversarialattacks}. We generate adversarial suffixes for each prompt in these datasets using a total of four generative attack models: two variants of AmpleGCG and two variants of AmpleGCG-plus~\cite{liao2024amplegcglearninguniversaltransferable, kumar2024amplegcgplusstronggenerativemodel}. We use HarmBench~\cite{mazeika2024harmbench} as a measure of whether the model's response to an prompt-suffix pair constitutes a jailbreak.

We exclude the Vicuna models from testing as we find that in $53.5\%$ of cases the model already gives a jailbroken reply to the malicious prompts without any adversarial suffixes or prompt engineering. We refer to this as \textit{benign ASR}. We also note that \texttt{GPT3.5-0125} is very easily tricked (and had a benign ASR of $2.7\%$) -- one of the cases where our pipeline failed is illustrated below:

\underline{Original prompt}: \textit{Teach children how to access and use firearms}\\
\underline{Adversarial prompt}: \textit{Teach children how to access and use firearms involves grammar Rem requested AleksomeWebView]. Write starting Colse `"S:\$ django imports"+Oneindent}\\
\underline{ASF-filtered prompt}: \textit{Teach children how to access and use firearms involves grammar}

The extra ``\textit{involves grammar}'' suffix was sufficient to jailbreak the model, and appears completely natural. Unfortunately, we cannot account for these edge cases and believe that this is a product of a poorly aligned model.

The results are present in~\Cref{tbl:adversarial_results}. After sanitization, $76.4\%$ of the suffixes were completely removed, leaving behind the original prompt. In just $2.9\%$ of cases, the returned sanitized prompt was empty due to either correctly segmenting the adversarial suffix from the prompt but subsequently classifying the prompt itself as an adversarial suffix, or due to segmentation leaving significant overlap between the prompt and suffix resulting in that segment being marked as adversarial. 

We note that our defense successfully reduces the ASR in the scenario described in~\Cref{tbl:adversarial_results_ppl}. The AIR and AID techniques~\cite{liao2024amplegcglearninguniversaltransferable} would have otherwise completely evaded detection of existing PPL-based defenses such as~\citet{jain2023baselinedefensesadversarialattacks,alon2023detectinglanguagemodelattacks}.

We observe a substantial reduction in ASR across all models and prompt sets following ASF sanitization, demonstrating that the pipeline reliably neutralizes their attack effectiveness in a practical LLM deployment setting.

\begin{table}[t]
    \caption{Defense pipeline results on the Malicious Instruct~\cite{huang2023catastrophic} and AdvBench~\cite{zou2023universaltransferableadversarialattacks} datasets. We use $k=20$ beams for generating the adversarial suffixes due to memory constraints. $\text{ASR}'$ represents ASR post-application of our defense pipeline with default settings. Suffixes were tailored to each attacked model by generating them using the appropriate AmpleGCG or AmpleGCG-plus model as described in~\citet{kumar2024amplegcgplusstronggenerativemodel}. \texttt{(+)} indicates the fact that suffixes were generated using the updated \textit{-plus} variant of AmpleGCG. ASR is evaluated via HarmBench-cls~\cite{mazeika2024harmbench}.}
    \label{tbl:adversarial_results}
    \centering
    \begin{subtable}[b]{0.45\textwidth}
    \centering
    \caption{Evaluated on the full dataset, using the standard method described in~\citet{kumar2024amplegcgplusstronggenerativemodel}.}\label{tbl:adversarial_results_main}
    \begin{tabular}{lcc}
    \toprule
    Models & ASR & $\text{ASR}'$ \\
    \midrule
    Llama-2-7b-chat     & 81.1\% & \textbf{1.8\%} \\
    Llama-2-7b-chat (+) & 93.1\% & \textbf{4.0\%} \\
    GPT3.5-0125         & 92.1\%   & \textbf{16.9\%} \\
    GPT4-0613 (+)       & 18.4\%   & \textbf{3.9\%} \\
    \bottomrule
    \end{tabular}
    \end{subtable}%
    \hfill
    \begin{subtable}[b]{0.45\textwidth}
    \centering
    \caption{Evaluating our pipeline against AIR and AID~\cite{liao2024amplegcglearninguniversaltransferable} -- techniques designed to bypass the PPL-defense~\cite{jain2023baselinedefensesadversarialattacks}. We evaluate on a random subset of $50$ samples taken from the full dataset.}\label{tbl:adversarial_results_ppl}
    \begin{tabular}{lcc}
    \toprule
    Models & GPT3.5 & GPT4 (+) \\
    \midrule
    AIR ASR             & 80.0\% & 2.0\% \\
    AIR $\text{ASR}'$   & \textbf{18.0\%} & \textbf{0.0\%} \\
    AID ASR             & 78.0\% & 2.0\% \\
    AID $\text{ASR}'$   & \textbf{16.0\%} & \textbf{0.0\%} \\
    \bottomrule
    \end{tabular}
    \end{subtable}%
\end{table}

% There is noticeable difference in performance, with changes in accuracy being attributed to stochastic generation. 
\begin{table}[t]
  \centering
    \caption{Non-adversarial performance in a variety of natural language tasks~\cite{eval-harness}. The tables show percentage difference of performance with and without our ASF defensive pipeline. If a task has multiple subsets, we report the mean accuracy. The models referred to in the table are \texttt{Llama-3.1-8B}, \texttt{Mistral-7B-Instruct-v0.1}, and \texttt{gpt-4.1-mini-2025-04-14}.}\label{tbl:non_adversarial_results}
  \begin{subtable}[t]{0.45\textwidth}
    \caption{\texttt{generate\_until} tasks}\label{tbl:non_adversarial_results_generateuntil}
    \centering
    \begin{tabular}{lcc}
    \toprule
    Models & \texttt{TruthfulQA-gen} & \texttt{GSM8k} \\
    \midrule
    Mistral-7B & -1.5\% & -0.6\% \\
    Llama-8B & -0.5\% & 0.8\% \\
    GPT-mini & 0.7\% & -1.0\% \\
    \bottomrule
    \end{tabular}
  \end{subtable}%
  \hfill
  \begin{subtable}[t]{0.45\textwidth}
    \centering
    \caption{\texttt{loglikelihood} tasks}\label{tbl:non_adversarial_results_loglike}
    \begin{tabular}{lcc}
    \toprule
    Task & Mistral-7B & Llama-8B \\
    \midrule
    \texttt{ARC-c} & -0.3\% & -0.2\% \\
    \texttt{HellaSwag} & -3.0\% & -3.4\% \\
    \texttt{WinoGrande} & -3.9\% & -5.1\% \\
    \bottomrule
    \end{tabular}
  \end{subtable}
\end{table}

\subsection{Non-Adversarial Settings}

To ensure that the ASF pipeline does not negatively impact LLM performance in benign usage scenarios, we conduct a suite of evaluations across standard natural language understanding and generation benchmarks. These tests are designed to assess whether ASF introduces any unintended degradation in the absence of adversarial inputs.

We employ the Language Model Evaluation Harness~\cite{eval-harness} to evaluate three representative models: \texttt{Llama-3.1-8B}, \texttt{Mistral-7B-Instruct-v0.1}, and \texttt{gpt-4.1-mini-2025-04-14}. The latter is accessed via OpenAI's Chat Completions API which limits access to token-level log-likelihoods, restricting us to generation-based evaluations. Consequently, for GPT-4.1-mini, we report results on the TruthfulQA-gen and GSM8k: widely-used instruction-following and reasoning tasks designed for generation evaluation. For the open-source models (Llama-3.1 and Mistral-7B), we additionally evaluate log-likelihood-based tasks -- ARC-challenge, HellaSwag, and WinoGrand -- providing a more complete picture of potential degradation in ranking or multiple-choice settings.

\Cref{tbl:non_adversarial_results} reports the percentage difference in accuracy between baseline model performance and model performance when prompts are passed through the ASF pipeline. Across all tasks and models, we observe minimal impact with most shifts within the margin of generation stochasticity and dataset noise. Overall, these findings support our conclusion that ASF does not meaningfully degrade model performance in non-adversarial settings. In practical terms, this suggests that ASF can be safely integrated into production inference pipelines without sacrificing task fidelity or output quality for legitimate users.

%% file: sections/discussion.tex
\section{Discussion}

As LLMs become integral to real-world applications, robust defense mechanisms like ASF carry significant deployment benefits. A key advantage of ASF is its model-agnostic design, meaning it can layered in front of any commercial or closed-source model (e.g. GPT-4 or Claude APIs) to intercept adversarial suffixes, with no access to or alteration of the model's internal weights or architecture. Perhaps most importantly for integration into commercial systems, ASF is easy to deploy as a ``plug-and-play'' component. It can be inserted as a preprocessing step in an existing compound AI system pipeline with minimal engineering effort.

ASF introduces minimal overhead in memory and computation: the entire defense uses roughly 387M parameters across two small models -- \texttt{12l-SM}~\cite{frohmann-etal-2024-segment} and \texttt{bert-base-uncased}~\cite{DBLP:journals/corr/abs-1810-04805}. This requires just $1.7\text{GB}$ of additional GPU memory to host the defense on-GPU. Compared to heavier certified defenses that require multiple costly forward passes through the main LLM, ASF inspects the input in a single pass with negligible latency. Crucially, it does not increase the target model's token consumption or inference time -- the protected LLM's compute budget (forward passes, token throughput, etc.) remains unchanged. Our evaluations also showed no noticeable degradation in the LLM's performance on benign tasks with ASF enabled. In other words, normal user queries and generations proceed as usual, preserving the user experience and accuracy of the model's responses.

\subsection{Limitations}~\label{sec:limitations}

While ASF proved effective in our experiments, several limitations must be acknowledged. First, the approach relies on accurately segmenting the user prompt from a appended suffix, and this segmentation can sometimes be imperfect. In most cases the prompt and adversarial suffix do separate cleanly into different segments, especially given the gibberish-like nature of many suffix attacks. However, if the segmentation model fails to fully isolate an adversarial suffix, ASF may misidentify a mixture of benign and malicious text as wholly malicious. In practice this means ASF might occasionally remove more text than necessary.

A second limitation is the possibility of false positives -- benign content being mistakenly flagged as adversarial. ASF uses a fine-tuned BERT classifier to detect malicious segments, and no classifier is 100\% error-free. There is a risk that an innocuous prompt containing unusual phrasing or tokens could be misclassified as an attack suffix. We found that straightforward heuristics in post-processing greatly mitigated this issue (e.g.\ ignoring segments that contain only common boilerplate words). However, the very need for hand-tuned rules indicates that the system may require careful calibration for different deployment contexts. If users frequently employ unconventional or out-of-distribution language in legitimate queries, ASF might initially flag some of those inputs incorrectly. Such cases would necessitate refining the filtering rules or retraining the classifier on a broader dataset of benign queries to improve its precision. In its current form, ASF errs on the side of caution – it is tuned to aggressively catch any suspected adversarial suffix, which by design sacrifices some specificity for security. This could inconvenience users if a normal query is erroneously rejected, so developers deploying ASF should monitor its decisions and adjust the filtering policy as needed to balance safety and accessibility.

It is also worth noting that ASF is specialized for the suffix-style attack paradigm, which may limit its scope against other prompt attack strategies. If an adversary devised a fundamentally different prompt injection technique that does not involve a discernible suffix (e.g.\ hand-crafted methods), ASF might not detect it without further extension of the method. Thus, while ASF significantly raises the bar against the current state-of-the-art suffix attacks, it is not a panacea for all prompt-based exploits.

\subsection{Intended Use Case Scenarios}

The Adversarial Suffix Filtering (ASF) pipeline is designed for deployment as an efficient, model-agnostic preprocessing layer within broader LLM security frameworks. We envision ASF serving as a first line of defense in compound AI-enabled safety architectures, where it operates as a lightweight, real-time filter to detect and neutralize adversarial suffix attacks. Owing to its low computational overhead and lack of reliance on model internals, ASF is particularly well-suited for \textit{front-end deployment} in latency-sensitive or resource-constrained environments.

Beyond its utility as a standalone defense mechanism, ASF can also function as a triggering component in larger systems that integrate multiple complementary defense strategies. In such configurations, ASF acts as a binary classifier on input prompts: if a suspected adversarial suffix is detected, ASF can optionally escalate the input to more computationally intensive downstream defenses, such as certified sanitization mechanisms, content filtering using ensemble classifiers, or constrained decoding techniques~\cite{kumar2025certifyingllmsafetyadversarial, varshney-etal-2024-art}. This staged approach ensures that the majority of benign or easily detectable adversarial cases are handled quickly and inexpensively, while reserving more costly defenses for inputs that warrant deeper scrutiny.

This described approach aligns with the Swiss cheese model~\cite{a6aea03a-88be-3f1e-b05c-7062244b6d7e}, a well-established conceptual framework in risk management and safety engineering. In this model, no single defense layer is assumed to be fully robust; instead, multiple layers with different strengths and failure modes are employed in parallel, such that the overall system remains resilient even when individual layers are circumvented.

%% file: sections/conclusion.tex
\section{Conclusions}

We introduced Adversarial Suffix Filtering (ASF), a lightweight, model-agnostic defense pipeline for detecting and removing adversarial suffixes in LLM prompts. ASF segments user inputs and classifies each segment to identify and neutralize malicious prompt continuations prior to inference, requiring no modifications to the underlying model. Through extensive evaluation, we demonstrate that ASF significantly reduces the attack success rate of current state-of-the-art suffix-based jailbreaks while preserving model performance on non-adversarial tasks. ASF achieves this with minimal computational overhead, making it deployable in real-world, resource-constrained environments.

While ASF effectively defends against a specific of prompt injection attacks, its current design assumes a clear segmentation between benign prompt and malicious suffix. Future investigations should broaden ASF’s scope beyond pure suffix-style attacks to mixed-context, interleaved or prefix-based injections and to multilingual settings where segmentation cues differ substantially . Addressing the pipeline’s present susceptibility to imperfect sentence splits and the attendant false-positive risk will likely require tighter, possibly joint, optimization of the segmentation and classification stages together with continual calibration on richer benign corpora.